\documentclass[11pt,a4paper]{article}
\usepackage[margin=1in]{geometry}
\usepackage[T1]{fontenc}
\usepackage[hyphens]{url}
\usepackage{graphicx}
\usepackage{natbib}
\usepackage{caption}
\usepackage{booktabs}
\usepackage{amsmath}
\usepackage{amssymb}
\usepackage{multirow}
\usepackage{subcaption}
\usepackage{bbding}
\usepackage{xcolor}
\usepackage{pifont}
\usepackage{tikz}
\usepackage{algorithm}
\usepackage{algorithmic}
\usepackage{enumitem}
\usepackage{pdflscape}
\usepackage{float}
\usepackage{adjustbox}
\usepackage{caption}
\usepackage{graphicx}
\usepackage{tabularx}
\usepackage{array}

\newcolumntype{C}{>{\centering\arraybackslash}X}
\newcommand{\MetricIcon}[2]{%
\tikz[baseline=(char.base)]{
\node[
    circle,
    fill=#1,
    inner sep=1.8pt
] (char) {\fontsize{9pt}{10.5pt}\selectfont\textcolor{black!78}{#2}};
}}
\newcommand{\DiceIcon}{\MetricIcon{black!8}{D}}
\newcommand{\JacIcon}{\MetricIcon{black!14}{J}}
\newcommand{\HDIcon}{\MetricIcon{black!20}{H}}
\newcommand{\ParamIcon}[2]{%
\tikz[baseline=(char.base)]{
\node[
    circle,
    draw=#1,
    fill=#1!10,
    line width=0.45pt,
    inner sep=0.7pt
] (char) {\footnotesize\ensuremath{\mathit{#2}}};
}}
\newcommand{\MagParam}{\ParamIcon{orange!85!black}{m}}
\newcommand{\NumParam}{\ParamIcon{cyan!70!blue}{n}}
\newcommand{\RangeParam}{\ParamIcon{magenta!75!black}{r}}

\newcommand{\GrayCross}{\textcolor{gray!45}{\ding{55}}}
\newcommand{\best}[1]{\textbf{#1}}
\newcommand{\second}[1]{\underline{#1}}

\title{Human-Inspired Asymmetric Collaboration for \\Dual-Student Semi-Supervised Medical Image Segmentation}
\author{Huiyi Ren \and Ziming Fang \and Jiaming Cai \\ Haojie Li \and Rui Xu}
\date{}

\begin{document}
\maketitle
\setlength{\textfloatsep}{6pt} 
\setlength{\floatsep}{5pt}
\setlength{\intextsep}{6pt} 

\setlength{\abovedisplayskip}{5pt plus 1pt minus 2pt}
\setlength{\belowdisplayskip}{5pt plus 1pt minus 2pt}
\setlength{\abovedisplayshortskip}{3pt plus 1pt}
\setlength{\belowdisplayshortskip}{3pt plus 1pt minus 1pt}

\begin{abstract}
Dual-student learning enriches predictive diversity in SSMIS by introducing a strongly augmented view alongside a weakly augmented view. Insufficient strong augmentation may homogenize the students, whereas excessive augmentation can corrupt anatomical evidence and destabilize collaboration.
This challenge echoes feedback-regulated human instruction, where learners receive differentiated tasks and the task difficulty is continually adjusted according to their evolving responses.
Inspired by these, we propose \emph{Human-Inspired Asymmetric Collaboration} (HAC), a closed-loop dual-student framework that maintains weak--strong learning through adaptive difficulty calibration and geometry preservation. Specifically, Reinforcement-guided Difficulty Calibration (RDC) formulates strong-augmentation control as a window-level reinforcement learning process, using augmentation history, realized perturbations, and training responses to dynamically adjust the augmentation magnitude, operation number, and magnitude range. Meanwhile, the Gramian Geometry Preservation Constraint (G2PC) preserves the early-stage three-branch geometry through global volume and pairwise area constraints, preventing shared supervision from homogenizing the two students. Comprehensive evaluations across LA, Pancreas, and BraTS-2019 validate both the segmentation effectiveness and capability-aware difficulty calibration of HAC. Code will be publicly available upon acceptance.
\end{abstract}

\section{Introduction}
Accurate medical image segmentation~\cite{jiao2024learning,xu2024advances} provides essential anatomical and pathological delineations for clinical diagnosis, treatment planning, and quantitative disease assessment. Despite the strong performance of deep segmentation networks~\cite{minaee2022image,tajbakhsh2020embracing}, their success largely depends on large-scale pixel- or voxel-level annotations. Acquiring such annotations~\cite{li2023pln} is particularly expensive in medical imaging, as the process requires substantial effort from experienced clinicians and becomes increasingly laborious for volumetric data. Meanwhile, the abundant unlabeled scans produced in clinical practice~\cite{cheplygina2019notsupervised} remain underutilized. Semi-supervised medical image segmentation (SSMIS) methods~\cite{bai2017semi,luo2021semi,xu2022allaround} therefore offer a practical solution by learning from a small labeled dataset together with a much larger unlabeled dataset.

The Teacher--Student paradigm~\cite{yu2019uncertainty,li2020transformation} is widely adopted in existing SSMIS methods, where an exponential moving average (EMA) Teacher generates relatively stable pseudo-labels for unlabeled images while a Student is optimized to maintain prediction consistency under input perturbations. However, the confirmation-bias problem identified in pseudo-label learning~\cite{arazo2020pseudo,liu2024diffrect} can arise because a single Student provides limited representational diversity: its prediction errors may be accumulated by the EMA Teacher and subsequently returned as erroneous supervision.

\begin{figure}[!t]
    \centering

    \includegraphics[
        width=\linewidth,
        keepaspectratio
    ]{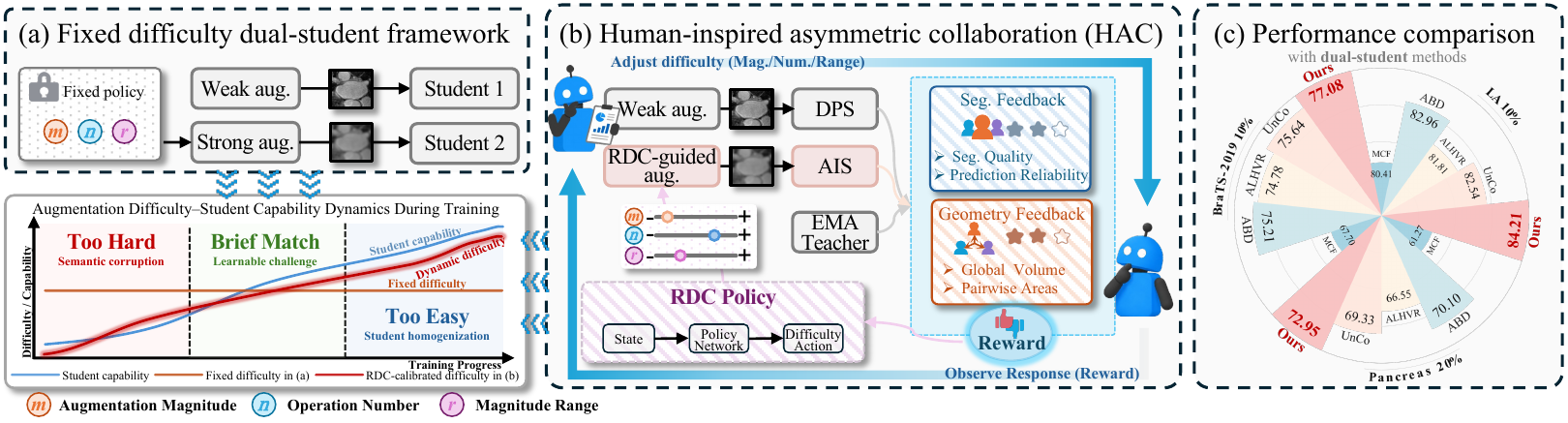}

    \caption{Motivation, core design, and performance of HAC.
    (a) A fixed augmentation difficulty becomes mismatched
    with evolving student capabilities.
    (b) HAC dynamically calibrates augmentation difficulty
    through a feedback-regulated loop.
    (c) HAC consistently outperforms representative
    dual-student methods across three benchmarks
    in terms of Jaccard.}

    \label{fig:motivation}
\end{figure}

Cross Teaching between CNN and Transformer~\cite{luo2022cross} introduces heterogeneous network structures, uncertainty-aware dual-network methods~\cite{peiris2023uncertainty} diversify predictive views, MC-Net+~\cite{wu2022mutual} exploits decoder disagreement, and MCF~\cite{wang2023mcf} performs mutual correction. These asymmetries allow the students to exchange diverse predictions and correct one another, alleviating the limitations of a single Student. The weak--strong view strategy~\cite{dong2025fdhcl} is a common way to encourage diversity: one student receives a weakly augmented image, whereas the other receives a strongly augmented counterpart. Existing methods increase view diversity through strong augmentations~\cite{dong2025fdhcl}, gradient-saliency-aware CutMix~\cite{jiang2024gradient}, bidirectional Copy--Paste~\cite{bai2023bidirectional}, and adaptive spatial displacement~\cite{chi2024adaptive}. However, these strategies typically rely on manually specified configurations, including a fixed augmentation magnitude, a fixed number of operations, or a fixed magnitude range. Static augmentation configurations cannot track the students' evolving capabilities. A fixed policy may exceed their limited perturbation tolerance early in training, obscuring anatomical evidence, match their capabilities only briefly, and later become too easy to sustain pathway differentiation, ultimately leading to student homogenization (Fig.~\ref{fig:motivation}). Therefore, \textbf{the key challenge is to dynamically calibrate augmentation difficulty according to the training state}, maintaining learnable yet discriminative perturbations that foster complementary pathways without compromising shared task semantics.

This challenge finds a close parallel in human instruction, where learning tasks are differentiated and their difficulty is continually recalibrated in response to learner feedback. Studies of human tutoring~\cite{bloom1984two,wood1976role} show that instructional support should adapt to the learner's current capability, while the challenge-point framework~\cite{guadagnoli2004challenge} indicates that learning is maximized when task difficulty matches the learner's evolving skill level.
From this perspective, effective dual-student collaboration requires not only differentiated learning pathways, but also augmentation difficulty that evolves with student capabilities, so as to sustain meaningful complementarity throughout training.

% \vspace{-0.6em}
Motivated by the observation above, we propose \emph{Human-Inspired Asymmetric Collaboration} (HAC), a closed-loop dual-student framework for SSMIS. HAC first establishes differentiated learning pathways: the Detail-Preserving Student (DPS) learns from weakly augmented views to retain fine anatomical cues, whereas the Appearance-Invariant Student (AIS) receives strong appearance perturbations to acquire robust semantic representations, with an EMA Teacher serving as a shared semantic anchor. Instead of relying on a fixed strong-augmentation policy, we introduce Reinforcement-guided Difficulty Calibration (RDC), a window-level RL mechanism that conditions its policy on the current training state and updates it using segmentation and geometry feedback to dynamically recalibrate the augmentation magnitude~\MagParam, operation number~\NumParam, and magnitude range~\RangeParam{} for the next training window. To prevent student homogenization, G2PC treats the early-stage Gramian geometry of DPS, AIS, and the Teacher as the target and preserves it through global volume and pairwise area constraints during subsequent training.
In this way, HAC closes the loop between asymmetric learning, RL-guided difficulty recalibration, and geometry-preserved branch individuality.

Our contributions are summarized as follows:

\begin{itemize}[
  topsep=-3pt,
  partopsep=0pt,
  itemsep=0pt,
  parsep=0pt
]
    \item We propose \emph{Human-Inspired Asymmetric Collaboration} (HAC), a closed-loop dual-student SSMIS framework that couples asymmetric learning pathways with feedback-regulated augmentation difficulty calibration.

    \item We formulate Reinforcement-guided Difficulty Calibration (RDC), which casts strong augmentation control as a window-level RL process and dynamically adjusts augmentation strength, number of operations, and magnitude range according to the evolving training state.

    \item We introduce a Gramian Geometry Preservation Constraint (G2PC) that preserves the early-stage Gramian geometry naturally formed among DPS, AIS, and the Teacher, preventing student homogenization while maintaining branch-specific response characteristics.

    % \item Experiments on LA, Pancreas, and BraTS-2019 benchmarks demonstrate the effectiveness of HAC, and ablation studies verify the contributions of asymmetric collaboration, adaptive augmentation calibration, and G2PC.
    \item Comprehensive experiments across LA, Pancreas, and BraTS-2019 validate the effectiveness of HAC, with extensive quantitative, qualitative, and mechanism analyses further demonstrating that RDC dynamically aligns augmentation difficulty with the students' evolving capabilities, while G2PC preserves student individuality.
\end{itemize}

\begin{figure}[!t]
    \centering
    \includegraphics[width=\textwidth]{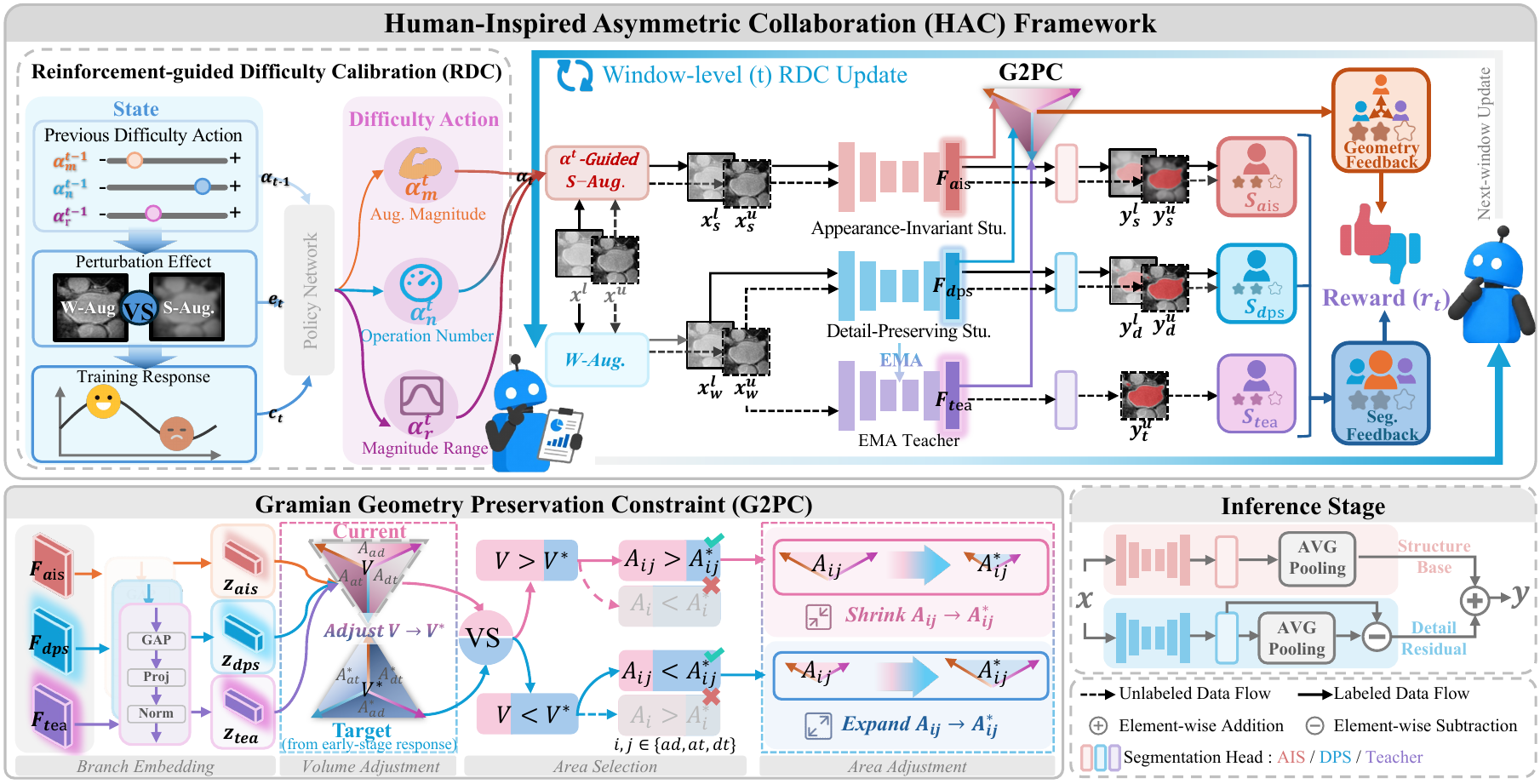}
    \vspace{-1.5em}
    \caption{Overview of HAC. After warm-up, RDC performs window-level updates to dynamically calibrate strong-augmentation difficulty from the evolving training state and feedback, while G2PC is applied at each iteration to preserve the target branch geometry through global volume and pairwise area constraints. For compactness, \(a\), \(d\), and \(t\) denote AIS, DPS, and the Teacher, respectively. During inference, predictions from the two complementary students are fused.}
    \label{fig:pipeline}
\end{figure}

% TODO: Add a Related Work section and citations after the reference list is available.
\section{Related Work}

\subsection{Semi-Supervised Medical Image Segmentation}

The single-student Teacher--Student paradigm~\cite{tarvainen2017mean,yu2019uncertainty,li2020transformation,luo2022semi} is widely used in semi-supervised medical image segmentation. Mean Teacher~\cite{tarvainen2017mean} updates the Teacher through an exponential moving average of the Student, providing temporally smoothed targets for unlabeled data, while UA-MT~\cite{yu2019uncertainty} and URPC~\cite{luo2022semi} further mitigate unreliable supervision through uncertainty-aware learning. However, as the Teacher is recursively updated from the Student, prediction errors may be preserved and progressively reinforced through iterative pseudo-label supervision. To alleviate this limitation, subsequent studies explore multi-branch collaborative learning. MC-Net+~\cite{wu2022mutual} leverages disagreement among heterogeneous decoders, MCF~\cite{wang2023mcf} performs mutual correction between networks, and AD-MT~\cite{zhao2024alternate} constructs diverse teaching signals. Nevertheless, most multi-branch methods induce branch differences through fixed architectural, supervisory, or input asymmetries, which cannot adapt to evolving model states and may become suboptimal as training progresses. HAC therefore dynamically adjusts branch specialization in response to the joint training state.

\subsection{Reinforcement Learning for Adaptive Training}

Data augmentation is essential for improving model generalization, yet its operation types and magnitudes are often manually specified. Automated augmentation methods therefore search for suitable policies according to downstream task performance. AutoAugment~\cite{cubuk2019autoaugment} uses reinforcement learning to search for augmentation policies, whereas Population Based Augmentation~\cite{ho2019population} learns schedules that evolve during training. Deep-RL augmentation for kidney-tumor segmentation~\cite{qin2020automatic} and adaptive sequence-length augmentation~\cite{xu2024automatic} extend this idea to medical image segmentation. These methods primarily optimize augmentation policies for a single target model and do not explicitly model how augmentation affects interactions among multiple collaborative branches. To address this limitation, HAC models the evolving joint state of multiple branches.

\section{Methodology}

% \subsection{Overview}

Given a labeled set $\mathcal{D}_l=\{(x_i^l,y_i^l)\}_{i=1}^{N}$ and an unlabeled set $\mathcal{D}_u=\{x_j^u\}_{j=1}^{M}$ with $N\ll M$, we propose \emph{Human-Inspired Asymmetric Collaboration} (HAC), a closed-loop dual-student framework that establishes two differentiated learning pathways, as shown in Fig.~\ref{fig:pipeline}. The Detail-Preserving Student (DPS) $\mathcal{F}_{dps}$ learns from weakly augmented views to retain fine anatomical cues, whereas the Appearance-Invariant Student (AIS) $\mathcal{F}_{ais}$ learns from strongly augmented views to acquire robust semantic representations, with an EMA Teacher $\mathcal{F}_{tea}$ providing stable pseudo-supervision.
To adapt strong-view difficulty to the students' evolving capabilities, we introduce Reinforcement-guided Difficulty Calibration (RDC), which recalibrates the augmentation configuration at the window level using segmentation and geometry feedback. G2PC preserves early-stage branch individuality to prevent student homogenization, forming a closed optimization loop summarized in Algorithm~\ref{alg:hac}.

\subsection{Reinforcement-guided Difficulty Calibration}

Although weak--strong augmentation pairs diversify dual-student learning, their fixed configurations cannot adapt to the students' evolving capabilities. RDC therefore formulates strong-augmentation calibration as a window-level reinforcement learning process with a \emph{state--action--feedback} loop, dynamically determining the current difficulty from the preceding-window state and using the resulting feedback to recalibrate it for the next window.

\noindent\textbf{State Construction.}
At the beginning of the $t$-th scheduling window, RDC constructs an eight-dimensional state $s_t$ from the previous difficulty action $\boldsymbol{\alpha}_{t-1}$, realized perturbation effect $e_t$, and model response $c_t$ of the preceding window:
\begin{equation}
s_t=[\boldsymbol{\alpha}_{t-1},e_t,c_t],\quad
\boldsymbol{\alpha}_{t-1}\in\mathbb{R}^{3},\
e_t\in\mathbb{R}^{2},\
c_t\in\mathbb{R}^{3}.
\label{eq:rdc-state}
\end{equation}
\ding{172} The previous difficulty action $\boldsymbol{\alpha}_{t-1}=[\alpha_m^{t-1},\alpha_n^{t-1},\alpha_r^{t-1}]$ records the augmentation magnitude, operation number, and magnitude range applied in the preceding window.
\ding{173} To quantify its realized effect, we compute each sample's mean weak-to-strong discrepancy $\delta_i=|\Omega|^{-1}\sum_{p\in\Omega}|x_w^i(p)-x_s^i(p)|$, and set $e_t=[\mu_\delta^{t-1},\sigma_\delta^{t-1}]$ using the mean and standard deviation over the preceding window.

\noindent\ding{174} The training-response term $c_t=[d_t,\Delta H_t,\rho_t]$ comprises the three-branch prediction discrepancy, AIS perturbation tolerance, and current training progress. Let $F_{dps}$, $F_{ais}$, and $F_{tea}$ denote their logits, and define $p_b=\operatorname{softmax}(F_b)$ for $b\in\mathcal{B}=\{dps,ais,tea\}$. Their prediction discrepancy is
\begin{equation}
d_t=
\frac{1}{6}
\sum_{b\neq b'}
D_{\rm KL}(p_b\|p_{b'}).
\end{equation}
The six KL divergences are averaged over the samples and voxels of the preceding window.
To estimate the perturbation tolerance of AIS, we partition each sample into high- and low-perturbation regions $\Omega_i^h$ and $\Omega_i^l$ according to $\left|x_w^i(p)-x_s^i(p)\right|$, and compute their window-averaged entropy gap:
\begin{equation}
\Delta H_t=\mathbb{E}_{i}\!\left[
\mathbb{E}_{p\in\Omega_i^h}H(p_{ais}^i(p))-
\mathbb{E}_{p\in\Omega_i^l}H(p_{ais}^i(p))
\right],
\end{equation}
where the inner expectations average the AIS prediction entropy over the high- and low-perturbation regions, while the outer expectation averages the resulting entropy gap over samples in the preceding window.
A larger $\Delta H_t$ indicates that AIS becomes substantially more uncertain in strongly perturbed regions, suggesting that the current augmentation difficulty is approaching or exceeding its tolerance. Finally, the normalized training progress is defined as $\rho_t=\tau_t/\tau_{\max}$, where $\tau_t$ is the global iteration at which $s_t$ is constructed and $\tau_{\max}$ is the total number of training iterations.

\begin{algorithm}[!t]
\caption{Modular Training Procedure of HAC}
% \vspace{-1.5em}
\label{alg:hac}
\small
\begin{algorithmic}[1]
\REQUIRE $\mathcal{F}_{dps},\mathcal{F}_{ais},\mathcal{F}_{tea}$; policy $\pi_\theta$; warm-up length $T_0$ and window length $K$
\STATE \textcolor{gray}{// Warm-up stage for $T_0$ iterations using fixed augmentation}
\STATE Optimize both students by $\mathcal{L}_{\rm warm}$ \hfill \textcolor{gray}{$\triangleright$ [Eq.~\ref{eq:warm-objective}]}
\STATE Estimate and fix $\mathcal{G}^*=(V^*,\{A_{ij}^*\})$ \hfill \textcolor{gray}{$\triangleright$ [Eq.~\ref{eq:g2pc-target}]}
\STATE \textcolor{gray}{// Fine-tune stage via feedback-regulated stage}
\FOR{each scheduling window $t$}
    \STATE Construct the state $s_t$ \hfill \textcolor{gray}{$\triangleright$ [Eq.~\ref{eq:rdc-state}]}
    \STATE Sample $\boldsymbol{\alpha}_t\sim\pi_\theta(\cdot\mid s_t)$ \hfill \textcolor{gray}{$\triangleright$ [Eq.~\ref{eq:rdc-action}]}
    \FOR{each mini-batch in the $K$-iteration window}
        \STATE Generate weak and $\boldsymbol{\alpha}_t$-guided strong views
        \STATE Compute $\mathcal{L}_{\rm G2PC}$ \hfill \textcolor{gray}{$\triangleright$ [Eq.~\ref{eq:g2pc-loss}]}
        \STATE Update both students by $\mathcal{L}_{\rm fine}$ \hfill \textcolor{gray}{$\triangleright$ [Eq.~\ref{eq:fine-objective}]}
\STATE Update $\mathcal{F}_{tea}$ from $\mathcal{F}_{dps}$ via EMA
    \ENDFOR
    \STATE Evaluate the feedback reward $r_t$ \hfill \textcolor{gray}{$\triangleright$ [Eq.~\ref{eq:rdc-reward}]}
    \STATE Update $\theta$ using $(s_t,\boldsymbol{\alpha}_t,r_t)$
\ENDFOR
\RETURN Students $\mathcal{F}_{dps}$ and $\mathcal{F}_{ais}$, EMA Teacher $\mathcal{F}_{tea}$
\end{algorithmic}
\end{algorithm}

\noindent\textbf{Difficulty Action Generation.}
Given the eight-dimensional state $s_t$, RDC employs a stochastic policy $\pi_\theta$ to determine the strong-augmentation difficulty for the t-th scheduling window. Following prior reinforcement-learning-based augmentation methods~\cite{cheng2019adversarial,yang2019searching}, we optimize $\pi_\theta$ using REINFORCE. Specifically, $\pi_\theta$ maps $s_t$ to a distribution over the three-dimensional difficulty action:
\begin{equation}
\boldsymbol{\alpha}_t=
[\alpha_m^t,\alpha_n^t,\alpha_r^t]
\sim\pi_\theta(\cdot\mid s_t),
\label{eq:rdc-action}
\end{equation}
where $\alpha_m^t$, $\alpha_n^t$, and $\alpha_r^t$ control the augmentation magnitude, operation number, and magnitude range, respectively. This configuration is held fixed throughout the window to generate strong views for AIS. After window training, segmentation and geometry feedback are converted into a reward to update $\pi_\theta$, enabling the policy to recalibrate the difficulty for the next window.

\noindent\textbf{Feedback-Guided Policy Update.}
At the end of window $t$, RDC evaluates $\boldsymbol{\alpha}_t$ using segmentation and geometry feedback. The former reflects the current segmentation quality and prediction reliability, while the latter measures branch homogenization through G2PC. They are integrated into a window-level reward:
\begin{equation}
r_t=r_t^{\rm seg}+\lambda_g r_t^{\rm geo},
\label{eq:rdc-reward}
\end{equation}
where $r_t^{\rm seg}$ and $r_t^{\rm geo}$ denote the window-level segmentation and geometry feedback scores, respectively, and $\lambda_g$ balances their contributions. Specifically, $r_t^{\rm seg}$ is the negative window average of the base training loss and Teacher prediction entropy, while $r_t^{\rm geo}$ is the negative window average of the G2PC loss. The resulting reward $r_t$ updates $\pi_\theta$, which generates the difficulty action for the next window from $s_{t+1}$.

Through such \emph{state--action--feedback} loop, RDC dynamically aligns strong-augmentation difficulty with the evolving training state, thereby promoting complementary learning between DPS and AIS.

\subsection{Gramian Geometry Preservation Constraint}

Shared supervision may gradually homogenize $\mathcal{F}_{dps}$ and $\mathcal{F}_{ais}$ despite their input asymmetry. To mitigate this tendency, we introduce G2PC to maintain their distinct representations throughout training.

\noindent\textbf{Target Geometry Construction.}
At the $t$-th warm-up iteration, $t=1,\ldots,T_0$, the decoder feature $F_b$ of branch $\mathcal{F}_b$, $b\in\mathcal{B}=\{ais,dps,tea\}$, is transformed by foreground-aware average pooling, a shared projection head, and $\ell_2$ normalization into $z_b^{(t)}$. The three branch embeddings form $\mathbf{Z}^{(t)}$, yielding $G^{(t)}=\mathbf{Z}^{(t)}\mathbf{Z}^{(t)\top}$ and the current Gramian volume $V$ and pairwise areas $A$:
\begin{equation}
V^{(t)}=\sqrt{\det(G^{(t)})},\quad
A_{ij}^{(t)}=\sqrt{G_{ii}^{(t)}G_{jj}^{(t)}-(G_{ij}^{(t)})^{2}},
\end{equation}
where $(i,j)\in\{(ais,dps),(ais,tea),(dps,tea)\}$. Rather than averaging the embeddings, G2PC averages these geometric measurements over the warm-up iterations:
\begin{equation}
\textstyle
V^*=\frac{1}{T_0}\sum_{t=1}^{T_0}V^{(t)},\quad
A_{ij}^*=\frac{1}{T_0}\sum_{t=1}^{T_0}A_{ij}^{(t)}
\label{eq:g2pc-target}
\end{equation}
Early geometry serves as the reference because it captures data-driven branch characteristics from asymmetric inputs before shared-supervision homogenization. Together, these statistics define $\mathcal{G}^*=(V^*,\{A_{ij}^*\})$, which guides the subsequent fine-tuning stage to prevent the two students from becoming homogeneous.

\noindent\textbf{Volume-Guided Area Preservation.}
G2PC first measures the global deviation as $\mathcal{L}_V=|V-V^*|$, and then uses its direction to selectively correct inconsistent pairwise areas:
\begin{equation}
\mathcal{L}_{A}=
\begin{cases}
|A_{ij}-A_{ij}^*|, & V>V^*\ \land\ A_{ij}>A_{ij}^*,\\
|A_{ij}-A_{ij}^*|, & V<V^*\ \land\ A_{ij}<A_{ij}^*,\\
0, & \text{otherwise}.
\end{cases}
\end{equation}
When $V>V^*$, only an area above its target is penalized and therefore shrunk; when $V<V^*$, only an area below its target is penalized and therefore expanded.

\noindent\textbf{Entropy-Adaptive Integration.}
The three pairwise area deviations may be either broadly distributed or concentrated in a particular branch relation. We use normalized entropy to quantify this dispersion and adaptively balance the two geometric constraints: high entropy favors global volume preservation, whereas low entropy emphasizes localized area correction. For $\delta_{ij}=A_{ij}-A_{ij}^*$, let $\bar{\delta}_{ij}=|\delta_{ij}|/\sum_{m<n}|\delta_{mn}|$ denote the normalized absolute deviation. Its entropy is
\begin{equation}
H_A=-\frac{1}{\log 3}
\sum_{i<j}\bar{\delta}_{ij}\log\bar{\delta}_{ij}.
\end{equation}
The final constraint is
\begin{equation}
\mathcal{L}_{\rm G2PC}
=H_A\mathcal{L}_{V}+\lambda_A(1-H_A)\mathcal{L}_{A}.
\label{eq:g2pc-loss}
\end{equation}
Here, $\lambda_A$ balances the two components, and $H_A$ is detached during optimization.

\subsection{Training Objective and Inference}

Training proceeds in two stages. During warm-up, a fixed augmentation configuration is adopted, and both students are optimized with conventional supervised objectives following Mean Teacher~\cite{tarvainen2017mean} and UA-MT~\cite{yu2019uncertainty}, collectively denoted by $\mathcal{L}_{\rm base}$ and comprising cross-entropy and Dice losses:
% During warm-up, a fixed augmentation configuration is adopted, and both students are optimized with conventional supervised and teacher-guided consistency objectives following Mean Teacher and UA-MT, collectively denoted by \(\mathcal{L}_{\rm base}\) and comprising cross-entropy and Dice losses:
\begin{equation}
\mathcal{L}_{\rm warm}=\mathcal{L}_{\rm base}.
\label{eq:warm-objective}
\end{equation}
The resulting early differentiated responses are used to construct the fixed target geometry $\mathcal{G}^*$.

During fine-tuning, HAC is optimized under the guidance of G2PC, with the training objective defined as:
\begin{equation}
\mathcal{L}_{\rm fine}=\mathcal{L}_{\rm base}+\lambda_{\rm G}\mathcal{L}_{\rm G2PC}.
\label{eq:fine-objective}
\end{equation}
G2PC is evaluated at every iteration to preserve branch individuality, while RDC recalibrates the strong-augmentation policy at the end of each window. During inference, predictions of DPS and AIS are fused as shown in Fig.~\ref{fig:pipeline}.

\section{Experiments}

\begin{table}[!t]
\centering

{\fontsize{8.5pt}{10.5pt}\selectfont
\setlength{\tabcolsep}{1pt}
\renewcommand{\arraystretch}{1.35}

\begin{tabular}{@{}l|ccc|ccc|ccc|ccc|ccc|ccc@{}}
\toprule
\midrule
\multirow[c]{3}{*}{\raisebox{-1.5ex}{Method}}
& \multicolumn{6}{c|}{LA}
& \multicolumn{6}{c|}{Pancreas}
& \multicolumn{6}{c}{BraTS-2019}
\\
\cmidrule(lr){2-7}\cmidrule(lr){8-13}\cmidrule(lr){14-19}
& \multicolumn{3}{c|}{5\% labeled data}
& \multicolumn{3}{c|}{10\% labeled data}
& \multicolumn{3}{c|}{10\% labeled data}
& \multicolumn{3}{c|}{20\% labeled data}
& \multicolumn{3}{c|}{10\% labeled data}
& \multicolumn{3}{c}{20\% labeled data}
\\
\cmidrule(lr){2-4}\cmidrule(lr){5-7}\cmidrule(lr){8-10}\cmidrule(lr){11-13}\cmidrule(lr){14-16}\cmidrule(lr){17-19}
& \DiceIcon $\uparrow$ & \JacIcon $\uparrow$ & \HDIcon $\downarrow$
& \DiceIcon $\uparrow$ & \JacIcon $\uparrow$ & \HDIcon $\downarrow$
& \DiceIcon $\uparrow$ & \JacIcon $\uparrow$ & \HDIcon $\downarrow$
& \DiceIcon $\uparrow$ & \JacIcon $\uparrow$ & \HDIcon $\downarrow$
& \DiceIcon $\uparrow$ & \JacIcon $\uparrow$ & \HDIcon $\downarrow$
& \DiceIcon $\uparrow$ & \JacIcon $\uparrow$ & \HDIcon $\downarrow$
\\
\midrule
VNet
& 62.36 & 53.27 & 36.41
& 80.24 & 68.82 & 17.65
& 56.23 & 41.30 & 20.19
& 71.93 & 58.59 & 12.24
& 75.25 & 65.92 & 16.94
& 80.39 & 69.10 & 13.01
\\

\midrule

% UA-MT~\cite{yu2019uncertainty}
UA-MT
& 82.26 & 70.98 & 13.71
& 86.28 & 76.11 & 18.71
& 66.44 & 52.02 & 17.04
& 76.10 & 62.62 & 10.84
& 81.31 & 70.38 & 14.68
& 82.88 & 72.70 & 11.58
\\

% URPC~\cite{luo2022semi}
URPC
& 82.48 & 70.18 & 14.65
& 86.92 & 77.03 & 11.13
& 73.53 & 59.44 & 22.57
& 80.02 & 67.30 & 8.51
& 81.24 & 70.72 & 10.90
& 82.14 & 70.75 & 10.00
\\

% MC-Net+~\cite{wu2022mutual}
MC-Net+
& 83.59 & 72.36 & 14.07
& 88.96 & 80.25 & 7.93
& 70.00 & 55.66 & 16.03
& 79.37 & 66.83 & 8.52
& 81.49 & 71.84 & 12.35
& 83.77 & 74.01 & 11.13
\\

% MCF~\cite{wang2023mcf}
MCF
& 87.06 & 77.83 & 7.81
& 88.71 & 80.41 & 6.32
& 67.71 & 53.83 & 17.17
& 75.00 & 61.27 & 11.59
& 77.89 & 67.70 & 11.67
& 81.95 & 72.11 & 11.22
\\

% BCP~\cite{bai2023bidirectional}
BCP
& 88.02 & 78.72 & 7.90
& 89.62 & 81.31 & 6.81
& 73.83 & 59.24 & 12.71
& 82.91 & 70.97 & 6.43
& 82.90 & 72.75 & 10.17
& 84.36 & 74.31 & 9.99
\\

% ABD~\cite{chi2024adaptive}
ABD
& 89.49 & 81.25 & 10.74
& 91.09 & 82.96 & 7.07
& 79.44 & 66.83 & 7.79
& 82.99 & 70.10 & 6.66
& 84.91 & 75.21 & 12.02
& 85.95 & 76.64 & 9.51
\\

% AD-MT~\cite{zhao2024alternate}
AD-MT
& 89.63 & 81.28 & \second{6.56}
& 90.55 & 82.79 & 5.81
& 80.21 & 67.51 & 7.18
& 82.61 & 70.70 & 4.94
& 85.50 & 75.88 & 9.32
& 86.31 & 77.01 & 8.42
\\

% ALHVR~\cite{lei2025adaptive}
ALHVR
& 87.72 & 78.25 & 7.99
& 89.95 & 81.81 & 7.65
& 75.62 & 62.72 & 11.57
& 78.99 & 66.55 & 8.17
& 84.53 & 74.78 & 13.00
& 85.08 & 75.66 & 8.58
\\

% UnCo~\cite{zeng2025uncertainty}
UnCo
& 88.83 & 80.02 & 6.99
& 90.37 & 82.54 & 6.11
& 78.53 & 65.20 & 7.36
& 81.49 & 69.33 & 6.35
& 85.09 & 75.64 & 8.63
& 85.43 & 76.33 & 9.05
\\

% BCSI~\cite{huang2026bidirectional}
BCSI
& 88.86 & 79.85 & 8.54
& 91.07 & 83.67 & 5.57
& \second{80.41} & \second{67.70} & \best{6.33}
& 83.24 & 71.61 & \second{4.87}
& \second{86.17} & \best{77.10} & \second{8.43}
& \second{86.86} & \second{77.80} & \second{7.62}
\\

% DualFete~\cite{yi2026dualfete}
DualFete
& \best{90.35} & \second{82.35} & \best{6.42}
& \second{91.28} & \second{83.81} & \second{5.51}
& 80.06 & 67.28 & \second{6.91}
& \second{83.86} & \second{72.43} & 5.19
& 85.82 & 76.77 & 8.80
& 85.37 & 76.27 & 8.63
\\

\midrule

Ours
& \second{90.28} & \best{82.36} & 6.89
& \best{91.41} & \best{84.21} & \best{5.46}
& \best{81.14} & \best{68.86} & 9.52
& \best{84.23} & \best{72.95} & \best{4.86}
& \best{86.20} & \second{77.08} & \best{7.96}
& \best{87.10} & \best{78.26} & \best{7.05}
\\

\midrule
\bottomrule
\end{tabular}
}

\caption{
Results on the LA, Pancreas, and BraTS-2019 datasets under different
labeled-data ratios. \protect\DiceIcon{}, \protect\JacIcon{}, and \protect\HDIcon{}
denote Dice, Jaccard, and 95HD, respectively. Higher Dice and Jaccard are better,
while lower 95HD is better. Best and second-best results are shown
in \textbf{bold} and \underline{underlined}, respectively.
}

\label{tab:main_comparison}
\end{table}

\begin{figure}[t]
    \centering
    \includegraphics[width=\columnwidth]{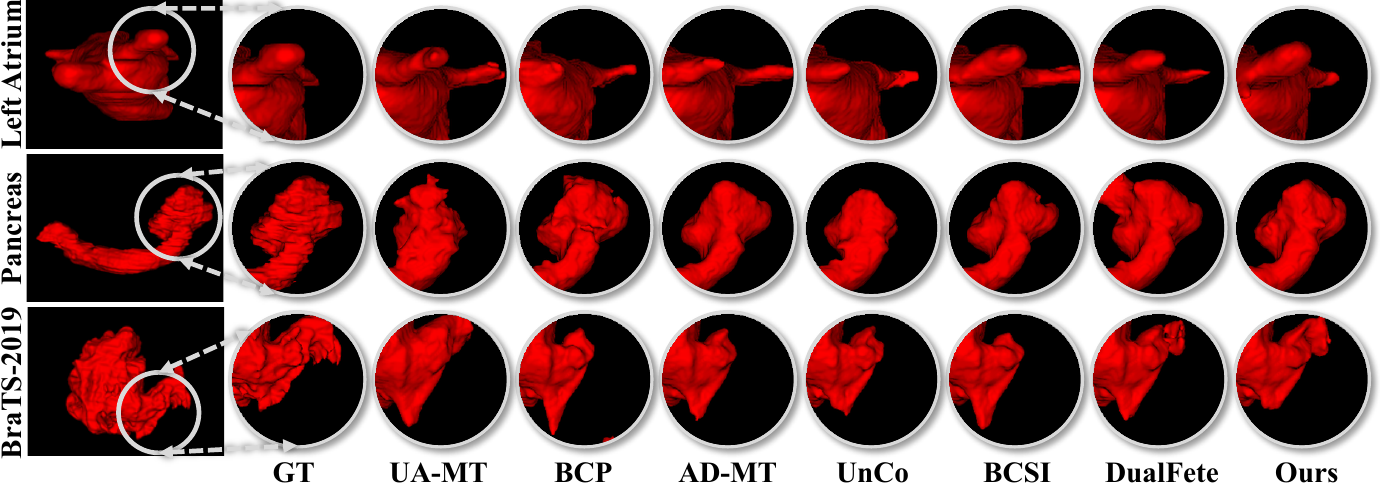}
    \vspace{-1.0em}
    \caption{Qualitative visualization of segmentation results.}
    
    \label{fig:vis1}
\end{figure}

\subsection{Experimental Settings}

\noindent\textbf{Datasets.}
Experiments are conducted on LA~\cite{xiong2021global}, Pancreas~\cite{roth2015deeporgan}, and BraTS-2019~\cite{menze2014multimodal}. Following DualFete~\cite{yi2026dualfete} and BCSI~\cite{huang2026bidirectional}, we use their data splits, labeled-subset selections, and preprocessing protocols.
\ding{172} LA contains 100 cardiac MRI scans split into 80 training and 20 testing cases, with 5\% and 10\% labeled settings.
\ding{173} Pancreas contains 82 abdominal CT scans split into 62 training and 20 testing cases, with 10\% and 20\% labeled settings.
\ding{174} BraTS-2019 contains 335 brain MRI scans split into 250 training, 25 validation, and 60 testing cases under the 10\% and 20\% labeled settings.

% TODO: Add dataset sizes, train/validation/test splits, preprocessing,
% augmentation ranges, backbone, optimizer, learning rate, training iterations,
% hardware, random seeds, and statistical reporting.

\noindent\textbf{Implementation Details.}
Our method is implemented in PyTorch and trained on a single 48-GB NVIDIA A40 GPU. Following prior work, we use V-Net as the backbone and re-run UnCo~\cite{zeng2025uncertainty} and DualFete~\cite{yi2026dualfete} on BraTS-2019 with the same backbone. We use SGD with a learning rate of 0.01, momentum of 0.9, weight decay of 0.0001, and a batch size of 4. Training lasts 15k iterations on LA and Pancreas and 30k on BraTS-2019. Following the staged warm-up practices of BCP~\cite{bai2023bidirectional} and BCSI~\cite{huang2026bidirectional}, $T_0$ is set to 2k iterations for LA and Pancreas and 4k for BraTS-2019, with the initial augmentation configuration fixed as $(\alpha_m,\alpha_n,\alpha_r)=(0.5,0.5,0.1)$. Input patches are randomly cropped to $112\times112\times80$ for LA and $96\times96\times96$ for the other datasets. Sliding-window inference uses strides of $(18,18,4)$ and $(16,16,16)$, respectively. We report Dice and Jaccard for regional overlap and the 95th-percentile Hausdorff distance (95HD) for boundary accuracy across all datasets and labeling ratios.

\subsection{Comparison with State-of-the-Art Methods}

Alongside the V-Net baseline, we compare HAC with 11 representative SSMIS methods: UA-MT~\cite{yu2019uncertainty}; URPC~\cite{luo2022semi} and MC-Net+~\cite{wu2022mutual}; MCF~\cite{wang2023mcf} and BCP~\cite{bai2023bidirectional}; ABD~\cite{chi2024adaptive}; AD-MT~\cite{zhao2024alternate}; ALHVR~\cite{lei2025adaptive}; UnCo~\cite{zeng2025uncertainty}; and BCSI~\cite{huang2026bidirectional} and DualFete~\cite{yi2026dualfete}.

\noindent\textbf{Quantitative comparison.}
As shown in Table~\ref{tab:main_comparison}, HAC achieves consistently strong performance on both regional and boundary metrics across all three datasets and labeled-data ratios. With 10\% labeled LA data, HAC outperforms DualFete by 0.13 percentage points in Dice. On Pancreas with 20\% labeled data, it improves Dice and Jaccard by 0.37 and 0.52 points and reduces 95HD by 0.33 over DualFete; on BraTS-2019 with 10\% labeled data, it reduces 95HD by 0.47 over BCSI. HAC also achieves the best overall performance on Pancreas with 10\% labeled data and BraTS-2019 with 20\% labeled data, demonstrating its robustness across imaging modalities, target structures, and annotation ratios.

\noindent\textbf{Qualitative comparison.}
% HAC better preserves fine-grained structures, recovers complete target regions, and suppresses false positives across all three datasets, with the enlarged regions further showing more accurate and continuous boundaries as shown in Fig.~\ref{fig:vis1}.
The qualitative results in Fig.~\ref{fig:vis1} further support the quantitative comparisons. Across the three datasets, HAC better preserves fine-grained anatomical structures, recovers more complete target regions, and suppresses spurious foreground predictions. Enlarged regions reveal more accurate and continuous boundaries, particularly in small, ambiguous regions, consistent with the Dice, Jaccard, and 95HD improvements in Table~\ref{tab:main_comparison}.

\begin{figure}[!t]
    \centering

    \includegraphics[
        width=0.95\linewidth,
        keepaspectratio
    ]{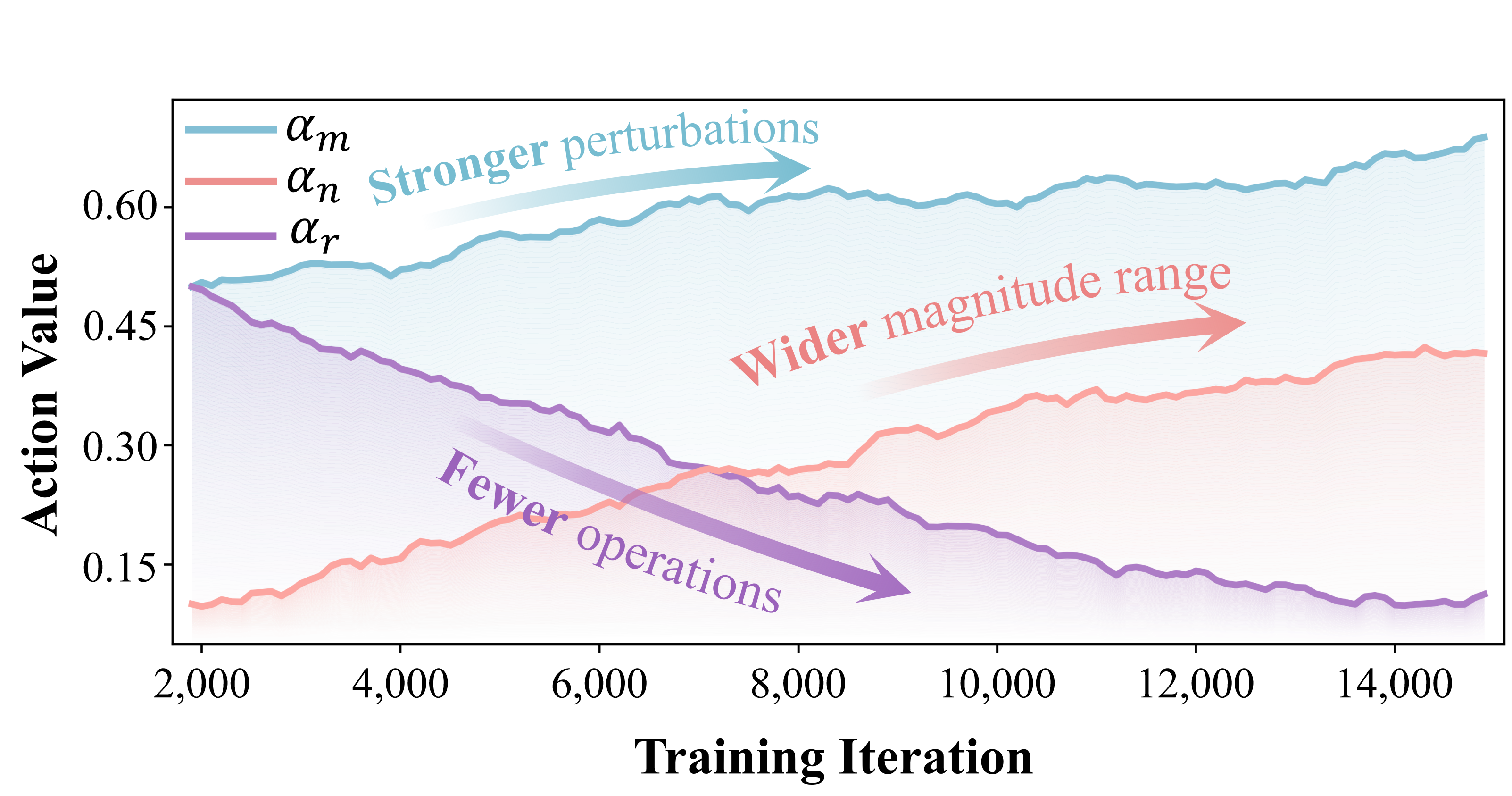}

    \caption{Training evolution of the RDC policy parameters.}
    \label{fig:parameter-evolution}
\end{figure}

\subsection{Mechanism Analysis of HAC}
\begin{figure}[!t]
    \centering
\includegraphics[width=\columnwidth]{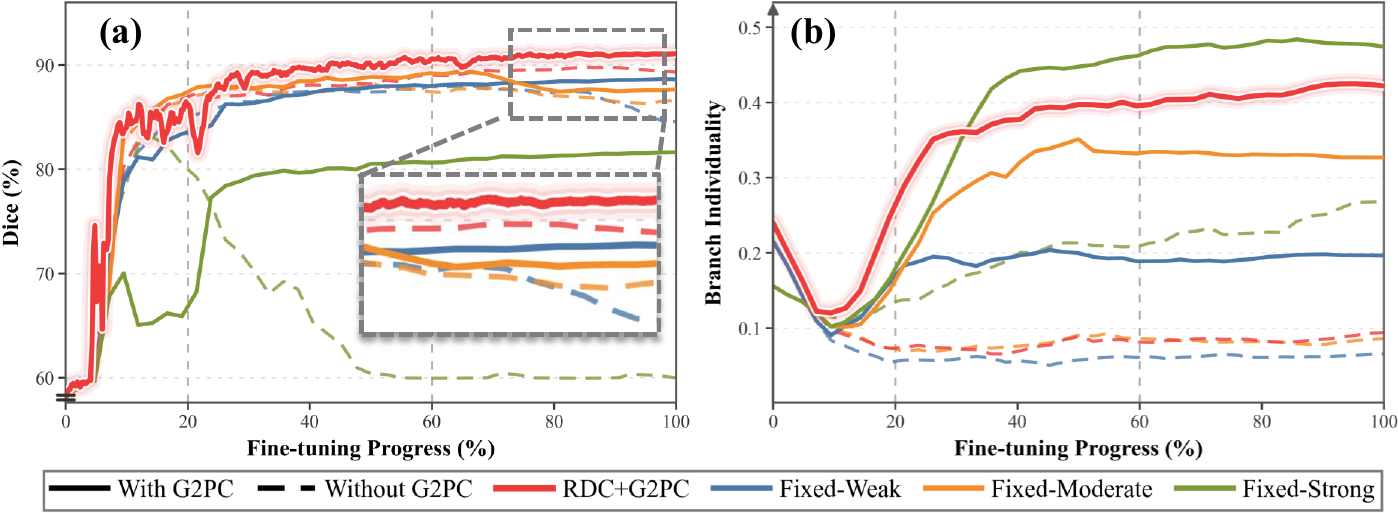}
\caption{Training dynamics of RDC and G2PC on LA with 10\% labeled data. (a)  Performance under fixed and adaptive augmentation. (b) Brach individuality  measured by cosine distance, with and without G2PC.}
    \label{fig:curve}
\end{figure}
\begin{figure}[!h]
    \centering
    \includegraphics[width=\columnwidth]{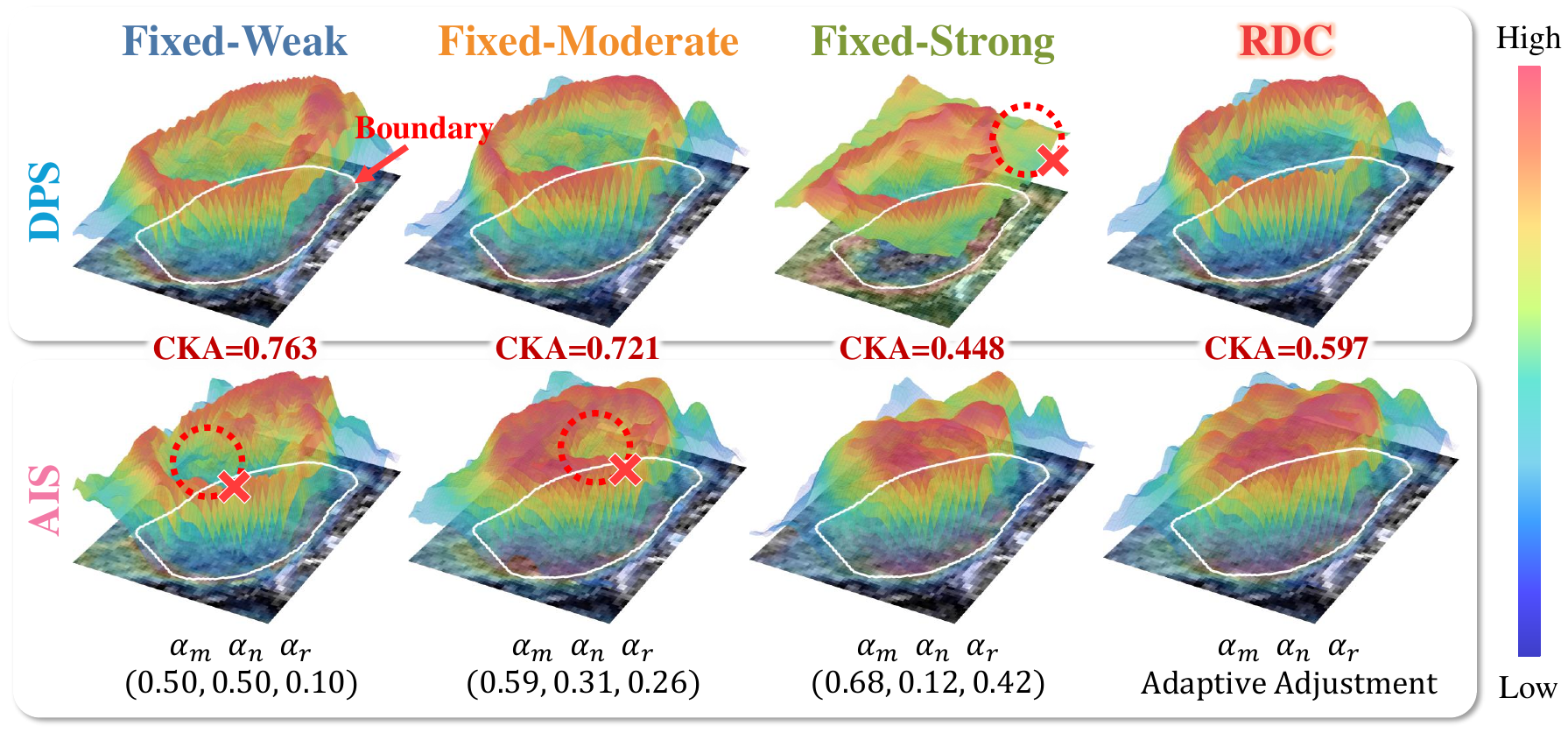}
    \caption{Feature responses and CKA similarity between DPS and AIS under different fixed and adaptive augmentations on LA with 10\% labeled data.}
    \label{fig:feat_vis2}
\end{figure}
\noindent\textbf{Why is adaptive augmentation necessary?}
No fixed augmentation difficulty remains suitable throughout training (Fig.~\ref{fig:curve}(a)): Fixed-Weak learns rapidly but later saturates, Fixed-Strong impedes early learning, and Fixed-Moderate provides only a suboptimal compromise. Specifically, Fixed-Weak uses the initial augmentation configuration, Fixed-Strong adopts the final policy learned by RDC, and Fixed-Moderate uses the element-wise averages of their three augmentation parameters. The corresponding parameter values are provided in Fig.~\ref{fig:feat_vis2}. The feature responses further show that weak augmentation induces highly similar representations, whereas strong augmentation corrupts task-relevant semantics; RDC instead preserves task-relevant information while maintaining distinct student representations (Fig.~\ref{fig:feat_vis2}). This advantage arises from its stage-wise calibration of augmentation magnitude, operation number, and magnitude range as shown in Fig.~\ref{fig:parameter-evolution}.

\noindent\textbf{Can G2PC preserve student individuality?}
Although adaptive augmentation differentiates the student inputs, shared pseudo-supervision can still homogenize their representations. We quantify student homogenization using Centered Kernel Alignment (CKA) between the foreground feature matrices of DPS and AIS, where a higher CKA indicates greater representational similarity~\cite{kornblith2019similarity}. Across fixed weak, moderate, and strong augmentations as well as RDC, this tendency intensifies without G2PC but is consistently suppressed when G2PC is applied (Fig.~\ref{fig:curve}(b)). By preserving the early-stage three-branch Gramian geometry, G2PC complements RDC by ensuring that the adaptively induced pathway differences remain distinct and meaningful throughout subsequent collaboration. Consequently, DPS emphasizes fine structures and boundaries, whereas AIS captures more complete target regions (Fig.~\ref{fig:feat_vis2}).

\subsection{Ablation and Sensitivity}

Since the main comparison covers three datasets and multiple labeled-data ratios, unless otherwise specified, all subsequent analyses are conducted on LA with 10\% labeled data to ensure consistent comparison. 

\noindent\textbf{Overall component analysis.}
As shown in Table~\ref{tab:overall-ablation}, all components contribute to the final performance. Compared with DPS alone, the dual-student baseline improves Dice by 3.83 percentage points, while reducing 95HD by 4.92 relative to AIS alone, confirming the complementarity of weak- and strong-view learning. Adding either G2PC or RDC improves both branches. With both modules, DPS reduces 95HD from 7.81 to 5.46, while AIS increases Dice from 89.59\% to 91.32\%. Their fused prediction achieves 91.41\% Dice and 5.46 95HD, outperforming the dual-student baseline by 1.80 percentage points in Dice and by 2.34 in 95HD. These results show that RDC and G2PC jointly reinforce the boundary-oriented role of DPS and the region-oriented role of AIS.

\begin{table}[!t]
\centering

% ========================================
% Left: Table 2 - Overall Ablation
% ========================================
\begin{minipage}[t]{0.52\linewidth}
\vspace{0pt}
\centering

{\fontsize{8.5pt}{10.5pt}\selectfont
\setlength{\tabcolsep}{1pt}
\renewcommand{\arraystretch}{1.2}

\begin{tabularx}{\linewidth}{@{}CCCC|CC|CC|CC@{}}
\toprule
\midrule

\multicolumn{4}{c|}{Methods}
& \multicolumn{2}{c|}{DPS}
& \multicolumn{2}{c|}{AIS}
& \multicolumn{2}{c}{Full} \\

\cmidrule(lr){1-4}
\cmidrule(lr){5-6}
\cmidrule(lr){7-8}
\cmidrule(lr){9-10}

DPS & AIS & G2PC & RDC
& \DiceIcon $\uparrow$
& \HDIcon $\downarrow$
& \DiceIcon $\uparrow$
& \HDIcon $\downarrow$
& \DiceIcon $\uparrow$
& \HDIcon $\downarrow$ \\

\midrule

\Checkmark & \GrayCross & \GrayCross & \GrayCross
& 85.78 & 13.04
& -- & --
& -- & -- \\

\GrayCross & \Checkmark & \GrayCross & \GrayCross
& -- & --
& 88.79 & 12.72
& -- & -- \\

\Checkmark & \Checkmark & \GrayCross & \GrayCross
& 88.94 & 7.81
& 89.59 & 7.91
& 89.61 & 7.80 \\

\Checkmark & \Checkmark & \Checkmark & \GrayCross
& 89.24 & 7.13
& 90.44 & 7.32
& 90.51 & 6.81 \\

\Checkmark & \Checkmark & \GrayCross & \Checkmark
& 89.35 & 6.40
& 90.58 & 7.17
& 90.56 & 6.54 \\

\Checkmark & \Checkmark & \Checkmark & \Checkmark
& 90.01 & 5.46
& 91.32 & 6.51
& \textbf{91.41} & \textbf{5.46} \\

\midrule
\bottomrule
\end{tabularx}
}

\captionsetup{font=footnotesize,skip=6pt}
\caption{Overall and branch-wise ablation results.
A dash indicates an inapplicable result for
single-student settings.}
\label{tab:overall-ablation}

\end{minipage}
\hfill
% ========================================
% Right: Table 3 - G2PC Ablation
% ========================================
\begin{minipage}[t]{0.47\linewidth}
\vspace{0pt}
\centering

{\fontsize{8.5pt}{10.5pt}\selectfont
\setlength{\tabcolsep}{3pt}
\renewcommand{\arraystretch}{1.65}

\begin{tabular}{@{}
>{\centering\arraybackslash}m{0.9cm}
>{\centering\arraybackslash}m{0.9cm}
>{\centering\arraybackslash}m{1.0cm}
|ccc@{}}
\toprule
\midrule

Volume & Area & \shortstack[c]{Entropy\\weight}
& \DiceIcon $\uparrow$
& \JacIcon $\uparrow$
& \HDIcon $\downarrow$ \\

\midrule

\GrayCross & \GrayCross & \GrayCross
& 90.56 & 82.87 & 6.54 \\

\Checkmark & \GrayCross & \GrayCross
& 90.92 & 83.41 & 5.54 \\

\GrayCross & \Checkmark & \GrayCross
& 90.88 & 83.34 & 5.78 \\

\Checkmark & \Checkmark & \GrayCross
& 91.05 & 83.61 & 5.52 \\

\Checkmark & \Checkmark & \Checkmark
& \textbf{91.41}
& \textbf{84.21}
& \textbf{5.46} \\

\midrule
\bottomrule
\end{tabular}
}

\captionsetup{font=footnotesize,skip=6pt}
\caption{Ablation study of G2PC components.}
\label{tab:g2pc-ablation}

\end{minipage}

\end{table}

\begin{figure}[t]
    \centering
    \vspace{0.6em}
    \includegraphics[width=\columnwidth]{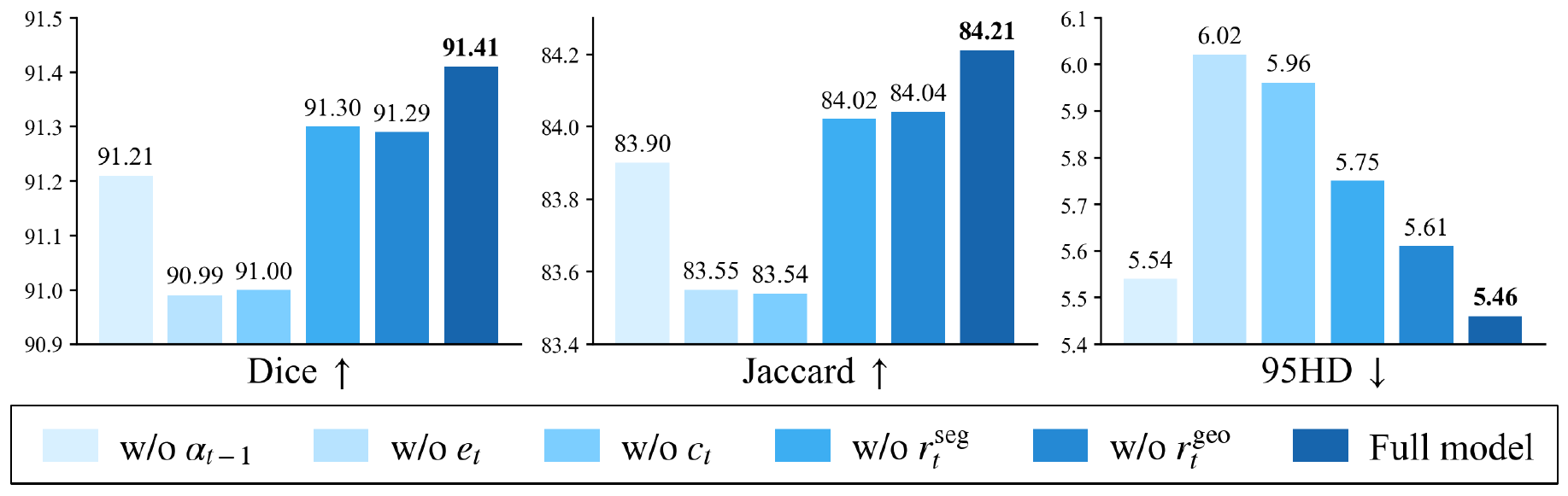}
    \vspace{-1.8em}
    \caption{Ablation study of the state components and reward terms in RDC.}
    \vspace{-0.6em}
    \label{fig:rl-ablation}
\end{figure}

\begin{figure}[!t]
    \centering

    % =================================
    % Left: 3D Ablation Visualization
    % =================================
    \begin{minipage}[t]{0.48\linewidth}
        \vspace{0pt}
        \centering

        \includegraphics[
            width=\linewidth,
            keepaspectratio
        ]{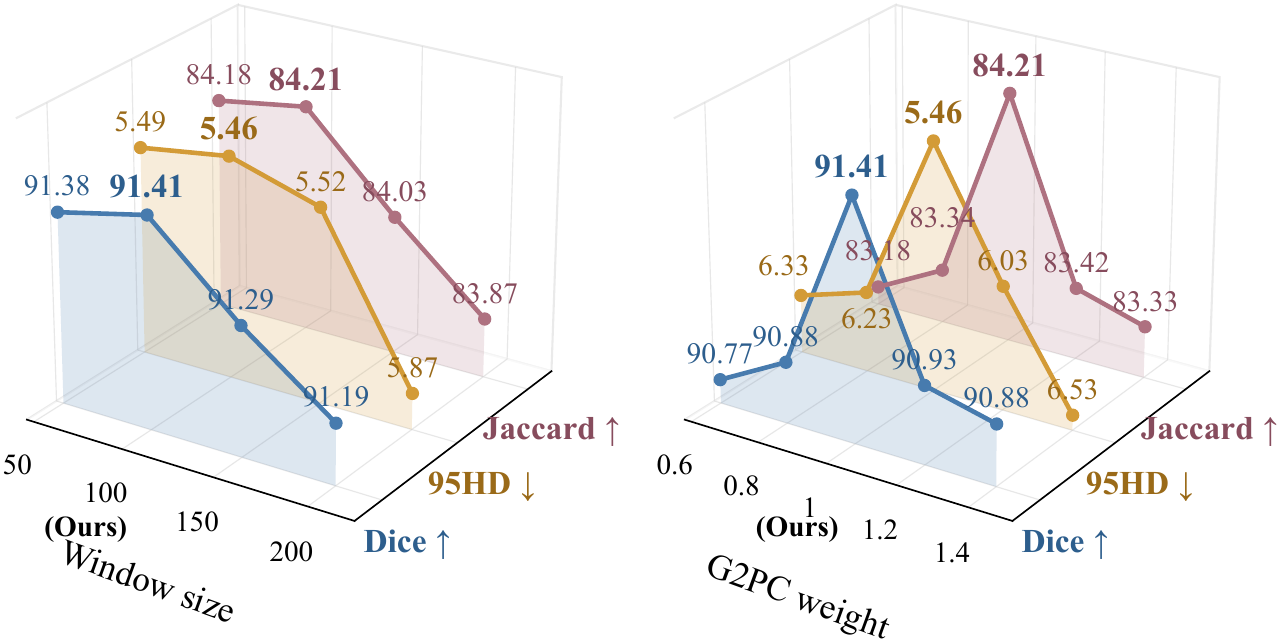}

        \captionsetup{font=footnotesize,skip=6pt}
        \caption{Visualization of the 3D ablation results.}
        \label{fig:ablation_3d_waterfall}
    \end{minipage}
    \hfill
    % =================================
    % Right: Performance-Efficiency
    % =================================
    \begin{minipage}[t]{0.48\linewidth}
        \vspace{0pt}
        \centering

        \includegraphics[
            width=\linewidth,
            keepaspectratio
        ]{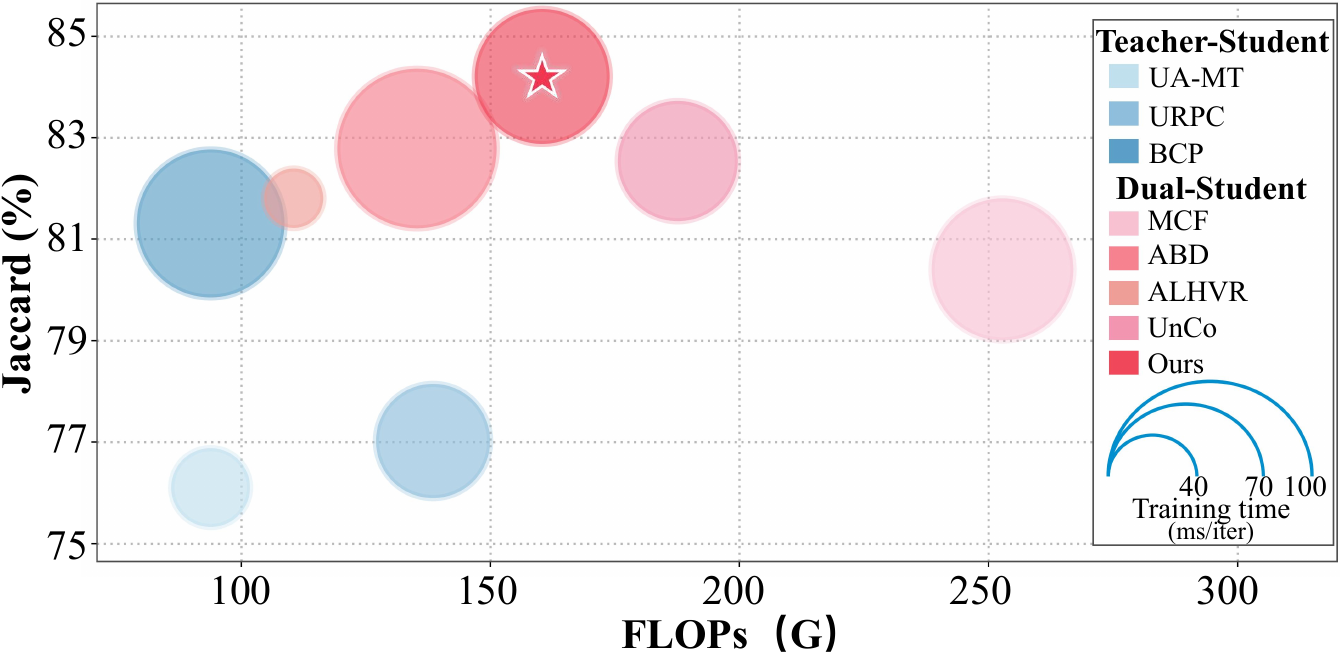}

        \captionsetup{font=footnotesize,skip=6pt}
        \caption{Performance--efficiency comparison.}
        \label{fig:tradeoff}
    \end{minipage}

\end{figure}

\noindent\textbf{G2PC component analysis.}
Table~\ref{tab:g2pc-ablation} further analyzes the internal components of G2PC. Either the global volume or pairwise area constraint improves all three metrics, showing that geometry preservation helps DPS and AIS retain their detail-preserving and appearance-invariant roles. Combining both raises Dice and Jaccard to 91.05\% and 83.61\%, while reducing 95HD to 5.52. With entropy-based weighting, the complete G2PC achieves the best results of 91.41\% Dice, 84.21\% Jaccard, and 5.46 95HD, validating the full design for maintaining complementary student specialization.

\noindent\textbf{RDC design analysis.}
For RDC, Fig.~\ref{fig:rl-ablation} evaluates its state and reward designs. Removing any state component degrades performance, confirming that the complete state is necessary for the RL policy to assess the current training condition and make appropriate decisions. In particular, excluding the perturbation effect $e_t$ or training response $c_t$ increases 95HD from 5.46 to 6.02 and 5.96, respectively. Removing either reward term also yields inferior results, whereas their combination achieves the best performance across all metrics.

\noindent\textbf{Hyperparameter sensitivity.}
For further validation, we vary the RDC window size \(K\) and G2PC weight \(\lambda_G\). Fig.~\ref{fig:ablation_3d_waterfall} demonstrates that \(K=100\) and \(\lambda_G=1.0\) achieve the best performance. Short windows provide insufficient samples for reliable reward estimation, whereas long windows reduce the frequency of feedback and policy updates, both hindering RL optimization. Similarly, a small \(\lambda_G\) weakens geometry preservation, while a large value may over-constrain task-oriented learning. The consistent performance across a broad parameter range further supports the robustness of the selected configuration.

\noindent\textbf{Efficiency.}
Dual-student collaboration introduces additional computation compared with Teacher--Student methods. We consider this trade-off favorable, as Fig.~\ref{fig:tradeoff} shows that HAC achieves the highest Jaccard among dual-student approaches while requiring fewer FLOPs than MCF and UnCo and a shorter per-iteration training time than ABD.

\section{Conclusion}

We present HAC, a human-inspired asymmetric collaboration framework for SSMIS that establishes differentiated learning pathways for DPS and AIS. RDC dynamically recalibrates augmentation magnitude, operation number, and magnitude range according to training feedback, while G2PC preserves the early-stage three-branch Gramian geometry to mitigate student homogenization and sustain complementary responses. Experiments on three benchmarks across diverse imaging modalities validate the effectiveness of HAC.

\bibliographystyle{plainnat}
\bibliography{references}

\end{document}